\documentclass{article}

\usepackage{PRIMEarxiv}

\usepackage[utf8]{inputenc} 
\usepackage[T1]{fontenc}    
\usepackage{hyperref}       
\usepackage{url}
\usepackage{enumitem}
\usepackage{booktabs}       
\usepackage{amsfonts}       
\usepackage{amsmath}
\usepackage{nicefrac}       
\usepackage{multirow}
\usepackage{microtype}      
\usepackage{lipsum}
\usepackage{fancyhdr}
\usepackage{color}
\usepackage{graphicx}       
\graphicspath{{media/}}     

\title{Learning Human Visual Attention on 3D Surfaces through Geometry-Queried Semantic Priors}

\author{%
Soham Pahari\textsuperscript{1} \quad
Sandeep C. Kumain\textsuperscript{1} \\[1.0ex]
\textsuperscript{1}UPES Dehradun \\[1.0ex]
\href{https://sohamrsch.github.io/semgeoattn}{sohamrsch.github.io/semgeoattn}
}

\begin{document}
\maketitle


\begin{abstract}

Human visual attention on three-dimensional objects emerges from the interplay between bottom-up geometric processing and top-down semantic recognition. Existing 3D saliency methods rely on hand-crafted geometric features or learning-based approaches that lack semantic awareness, failing to explain why humans fixate on semantically meaningful but geometrically unremarkable regions. We introduce SemGeo-AttentionNet, a dual-stream architecture that explicitly formalizes this dichotomy through asymmetric cross-modal fusion, leveraging diffusion-based semantic priors from geometry-conditioned multi-view rendering and point cloud transformers for geometric processing. Cross-attention ensures geometric features query semantic content, enabling bottom-up distinctiveness to guide top-down retrieval. We extend our framework to temporal scanpath generation through reinforcement learning, introducing the first formulation respecting 3D mesh topology with inhibition-of-return dynamics. Evaluation on SAL3D, NUS3D and 3DVA datasets demonstrates substantial improvements, validating how cognitively motivated architectures effectively model human visual attention on three-dimensional surfaces.
code: \url{https://github.com/sohampahari/SemGeoAttnNet}

\end{abstract}

\keywords{Visual attention modeling \and 3D saliency prediction \and Geometry-semantics fusion \and Diffusion-based semantic priors}

\section{Introduction}
\label{sec:intro}

Our real world belongs to the three-dimensional spaces. When people look at 3D objects, their attention comes from two processes working together: quick, automatic reactions to visual features (bottom-up processing) and thoughtful recognition based on knowledge (top-down processing). When we observe a 3D scene, low-level geometric features, edges, protrusions, surface discontinuities, trigger rapid, reflexive orienting responses, while high-level semantic understanding, e.g. recognizing faces, identifying functional parts, interpreting contextual meaning guides sustained, goal-directed fixations. Classical approaches to modeling 3D visual saliency have predominantly relied on hand-crafted geometric operators such as Gaussian curvature, mesh    saliency, or local rarity measures\cite{lee2005mesh, leifman2016surface, song2014mesh, tasse2015cluster}, operating under the assumption that attention is driven primarily by geometric distinctiveness. While these methods successfully identify geometrically conspicuous regions, they fundamentally fail to explain why humans fixate on semantically meaningful but geometrically unremarkable areas  a character's eyes command attention despite being shallow indentations, text on surfaces dominates gaze even when nearly flat, and worn tool handles attract fixations through recognition of function rather than geometric complexity. The critical limitation is that geometry alone cannot capture the semantic priors that fundamentally shape human visual exploration, and Martin el.\cite{martin2024sal3d} suffer from limited expressive capabilities, as they do not model the semantic and top-down cues.

Recent advances in deep learning have enabled data-driven attention models trained on eye-tracking datasets collected in virtual reality environments\cite{nousias2020mesh, song2021mesh, liu2023attention, zhou2022edvam}, offering more naturalistic viewing conditions than traditional laboratory setups\cite{wang2018tracking, lavoue2018visual}. However, existing neural approaches remain constrained by their reliance on geometric point cloud encoders that lack semantic awareness\cite{qi2017pointnet++}, or conversely, by 2D semantic models that fail to respect three-dimensional surface structure\cite{abid2020towards}. No prior work has explicitly formalized the bottom-up versus top-down dichotomy within the architecture itself.

We introduce SemGeo-AttentionNet, the first 3D visual attention model to explicitly bridge geometry and semantics through asymmetric cross-modal fusion. Building upon recent advances in diffusion-based semantic feature extraction for 3D surfaces\cite{dutt2024diffusion, rombach2022high, tang2023emergent}, we leverage these pretrained representations as frozen perceptual priors that provide zero-shot semantic understanding\cite{abdelreheem2023zero}. Our dual-stream architecture processes geometry through Point Transformer V3\cite{wu2024point} to capture multi-scale spatial structure, while compressing high-dimensional semantic features into a shared latent space. The fusion mechanism implements cross-attention where geometric features query semantic content, formalizing the principle that bottom-up distinctiveness guides top-down retrieval. This asymmetry ensures that even semantically important regions must exhibit geometric salience to attract attention, preventing collapse into pure semantic recognition while avoiding the limitations of pure geometric approaches.

Beyond static saliency prediction, we extend our framework to generate temporal scanpaths, sequences of fixations that capture the dynamics of visual exploration over time. Existing scanpath models operate exclusively on 2D images with pixel-based action spaces\cite{le2010overt}; We introduce the first formulation that respects 3D mesh topology, where actions navigate surface connectivity rather than Euclidean grids. We frame this as a partially observable Markov decision process with a reward function balancing saliency-driven exploitation against inhibition-of-return exploration, trained via proximal policy optimization\cite{schulman2017proximal}.

We evaluate our approach on SAL3D and NUS3D datasets, demonstrating substantial improvements over prior methods in both distributional accuracy and spatial structure preservation. Our contributions are threefold: 
\begin{itemize}[noitemsep, topsep=0pt]
    \item Diffusion-distilled semantic priors enable zero-shot semantic grounding in 3D attention modeling through asymmetric architectural fusion.
    \item Bottom-up and top-down attention mechanisms are formalized through explicit geometry-to-semantics cross-attention.
    \item A novel reinforcement learning framework for scanpath generation that respects mesh topology is presented.
\end{itemize}

This work establishes a new paradigm for modeling human visual attention on three-dimensional surfaces by explicitly unifying geometric and semantic processing within a cognitively motivated architecture.

This article follows a simple structure. We begin with a review of related work in section \ref{sec:related}, describe our proposed method in section \ref{sec:method}, present experiments and results in section \ref{sec:experiments}, and conclude in section \ref{sec:conclusion}.
\section{Related Work}
\label{sec:related}

\subsection{Hand-crafted Geometric Methods}

The foundation of 3D mesh saliency was established by \cite{lee2005mesh}, who introduced Gaussian-weighted mean curvature as the primary determinant of visual attention on geometric surfaces. This seminal work employed a center-surround operation to compute multi-scale saliency maps through non-linear suppression operators. Subsequent geometric approaches refined feature descriptors by incorporating Shannon entropy of mean curvatures \cite{page2003shape} to quantify local neighborhoods \cite{limper2016mesh}, and spatial regularization of vertices based on distance metrics \cite{leifman2016surface}. Spectral analysis methods based on the Laplace-Beltrami Operator emerged as a prominent direction, with \cite{song20133d, song2014mesh, song2018local} exploring eigenfunctions to capture intrinsic geometric properties. To mitigate noise sensitivity in high-frequency components, \cite{song2014mesh} employed log-Laplacian operators that preserve global structural characteristics. Clustering methods within the low-frequency domain were introduced by \cite{niu2020three}, while \cite{arvanitis2020robust} focused on detecting sharp features in industrial models. More recent geometric methods include \cite{dos2023saliency}, which proposed sub-sampling and interpolation for large-scale meshes, and \cite{wang2015multi}, which introduced low-rank analysis to separate novel features from common structures. Colorimetric principles were applied to colored meshes by \cite{nouri2024fully}, predicting visual saliency through vertex color analysis. Despite their efficiency, these hand-crafted methods rely solely on surface geometry and cannot model semantic recognition or top-down cognitive effects.

\subsection{Learning-based Approaches}

Deep learning introduced data-driven paradigms for 3D saliency prediction. Early neural approaches such as \cite{song2018local} combined global and local geometric features by integrating average and maximum Laplacian values through convolutional networks. The effectiveness of neural architectures on large mesh structures was demonstrated by \cite{nousias2020mesh}, building on earlier work in \cite{song2019mesh}. A major class of learning-based methods adopted 2D-to-3D projection strategies. \cite{song2021mesh} used weak supervision by training on image saliency datasets, predicting saliency for multiple viewpoints before projecting results back onto 3D meshes. Similarly, \cite{abid2020towards} relied on 2D saliency models trained on SALICON, and incorporated view-based rendering constraints. Point cloud segmentation networks also influenced several approaches. \cite{martin2024sal3d} adapted classification backbones to saliency prediction. These methods mainly rely on geometric encoders and treat saliency as a spatial problem. The importance of large-scale training data was highlighted by \cite{song20233d}, which showed improved performance with increased 2D supervision. Current neural models lack explicit semantic reasoning and fail to capture object-level importance that guides human attention.

\subsection{Texture and Semantic Integration}

Limitations of geometry-only methods motivated the integration of texture and semantic cues. Recent work on textured meshes by \cite{zhang2025mesh} and \cite{zhang2025textured} constructed datasets and models that combine 2D texture information with 3D geometry through UV mapping. Results show that color and geometry jointly influence visual attention. The MeshMamba architecture \cite{dao2024transformers, gu2024mamba} introduces graph convolution encoders that map mesh faces to UV pixel space. State space models such as Mamba have shown strong performance in vision tasks including image processing \cite{huang2024localmamba, zhu2024vision}, generation \cite{hu2024zigma, teng2024dim}, restoration \cite{guo2024mambair, zheng2024u}, and semantic segmentation \cite{pei2025efficientvmamba, ruan2024vm}. For point clouds, \cite{han2024mamba3d, liang2024pointmamba} leveraged Mamba for efficient processing. These texture-based methods require explicit UV textures created by artists or captured via scanning, limiting their use on untextured meshes. UV textures encode appearance but lack semantic understanding from large-scale vision-language models. Our method instead uses diffusion-based semantic priors, enabling zero-shot object-level recognition on raw meshes through geometry-conditioned multi-view rendering.

\subsection{Temporal Attention and Scanpath Modeling}

Beyond static saliency, temporal dynamics have been studied through scanpath analysis and task-driven attention. Task effects show clear differences between free viewing and goal-oriented behavior. \cite{le2010overt} compared free viewing with quality assessment in videos and observed similar priority maps with task-specific variations. \cite{liu2011visual} and \cite{alers2015effects} reported similar findings for image quality assessment. In immersive environments, \cite{hadnett2019effect} analyzed attention across viewing, search, and navigation in VR. \cite{hu2021ehtask} studied eye and head coordination in 360-degree videos. \cite{malpica2023task} examined exploration, memory, and search tasks in immersive scenes. For point clouds, \cite{reimat2021cwipc} conducted task-dependent eye tracking in VR. \cite{nguyen2024compeq} proposed task-free data collection in AR. \cite{zhou2023qava} studied quality assessment of dynamic point clouds. These studies show that high-level tasks strongly influence attention. However, scanpath generation remains limited to 2D images. No prior work models scanpaths directly on 3D mesh surfaces with actions respecting surface topology and 3D inhibition-of-return.

\section{Methodology}
\label{sec:method}
\begin{figure*}[htbp]
  \centering
  \includegraphics[width=\textwidth]{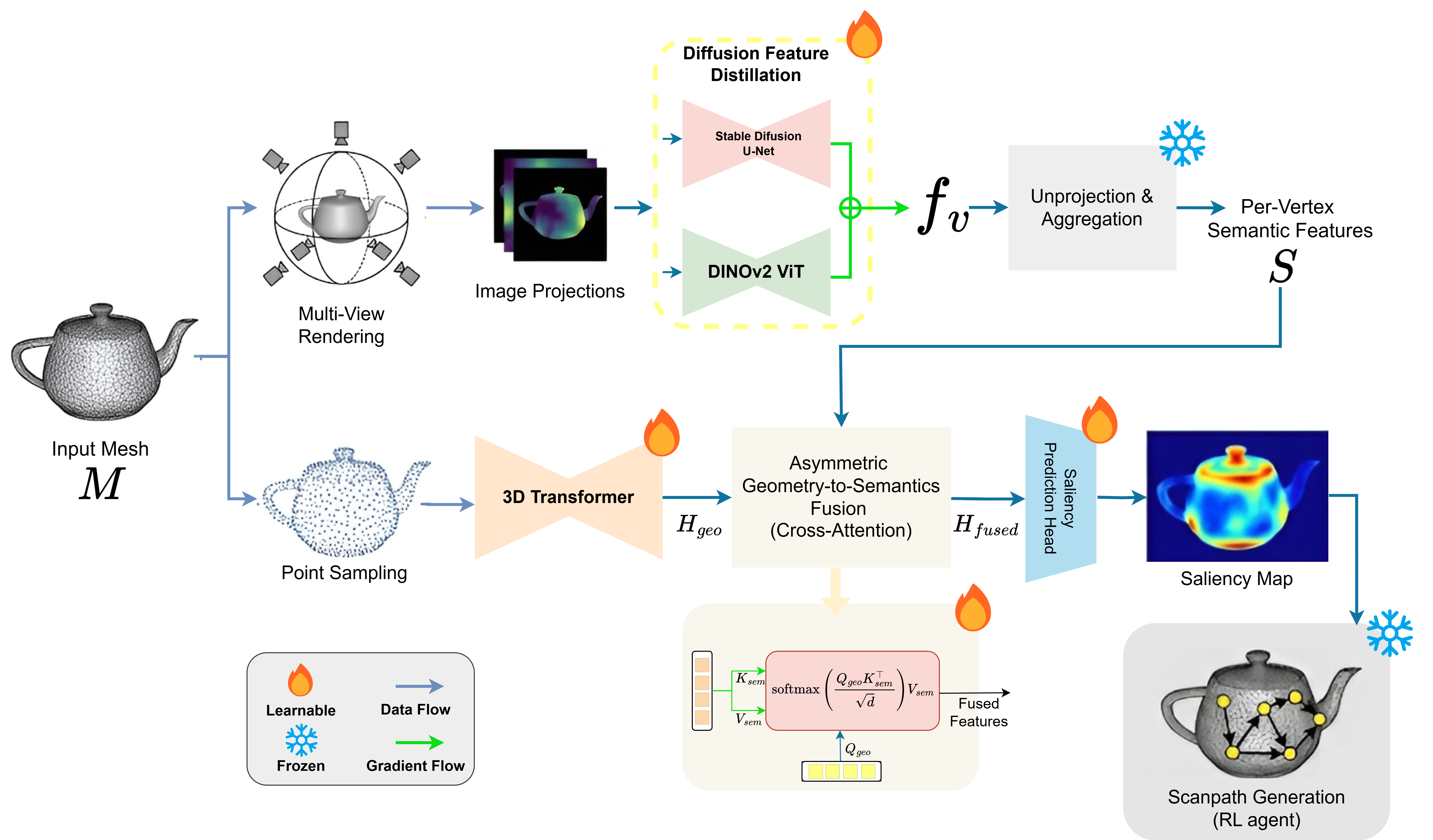} 
  \caption{\textbf{Method Overview.} We render the input mesh from multiple viewpoints and extract semantic features from geometry-conditioned Stable Diffusion U-Net (1280-dim) and DINOv2 (768-dim). These features are unprojected to mesh vertices, yielding 2048-dim semantic descriptors that are fused with Point Transformer V3 geometric features through cross-attention to predict per-vertex saliency.}
  \label{fig:pipeline}
\end{figure*}

Our approach addresses the fundamental challenge of predicting visual saliency on three-dimensional surfaces by explicitly modeling the interplay between bottom-up geometric distinctiveness and top-down semantic recognition. The central hypothesis is that human visual attention on 3D objects cannot be explained by geometry alone—flat regions like computer screens or faces attract gaze despite minimal curvature, while geometrically complex but semantically uninteresting areas (like noisy wall textures) are ignored. We propose SemGeo-AttentionNet (described in ~\ref{fig:pipeline}), a dual-stream architecture that treats geometry as a query mechanism and semantics as a contextual prior, enabling the model to adaptively weight these signals. We further extend this static prediction framework to generate temporal scanpaths through reinforcement learning, simulating sequential human exploration with inhibition of return on mesh surfaces.

\subsection{Overview and Attention Hypothesis} 

Let $\mathcal{M} = \{\mathbf{V}, \mathbf{F}\}$ denote a triangulated mesh with vertices $\mathbf{V} = \{\mathbf{v}_1, \ldots, \mathbf{v}_N\} \subset \mathbb{R}^3$ and faces $\mathbf{F}$. Our goal is to learn a mapping $f: \mathbf{V} \rightarrow [0, 1]^N$ that predicts per-vertex saliency scores $\hat{\mathbf{y}} = \{\hat{y}_1, \ldots, \hat{y}_N\}$, normalized such that $\sum_{i=1}^{N} \hat{y}_i = 1$, forming a probability distribution over fixation likelihood. Human visual attention is driven by two complementary mechanisms: bottom-up processing responds to local geometric anomalies (edges, protrusions), while top-down processing prioritizes semantically meaningful regions regardless of geometry. Our architecture mirrors this dichotomy—a geometric stream encodes spatial structure, a semantic stream captures object-level recognition, and an asymmetric fusion module allows geometry to query semantics, reflecting how low-level features guide retrieval of high-level meaning.

\subsection{Semantic Prior via Diffusion Feature Distillation} 

Classical geometric descriptors fail to capture semantic importance because they operate purely on surface properties. A flat television screen and a blank wall are geometrically identical but semantically distinct. To inject semantic awareness without requiring labeled 3D data, we leverage pretrained text-to-image diffusion models as a source of rich semantic priors. These models, trained on billions of image-text pairs, develop internal representations that distinguish semantic categories even when geometric cues are ambiguous.

We render the mesh $\mathcal{M}$ from $V = 100$ uniformly sampled viewpoints on a surrounding sphere, producing images $\{\mathbf{I}_v\}_{v=1}^{V}$ via a projection operator $\pi_v: \mathcal{M} \rightarrow \mathbb{R}^{H \times W \times 3}$. For each view, we extract geometric conditioning maps—a depth map $\mathbf{D}_v \in \mathbb{R}^{H \times W}$ and a surface normal map $\mathbf{N}_v \in \mathbb{R}^{H \times W \times 3}$—that guide a conditional diffusion inpainting process. Using ControlNet-conditioned Stable Diffusion with a text prompt $\tau$ describing the object category, we perform denoising from timestep $T$ to $1$, extracting intermediate U-Net decoder activations $\mathbf{F}_v^{(t)} \in \mathbb{R}^{H' \times W' \times D_{\text{diff}}}$ at each step $t$. These features capture both coarse semantic structure (early noisy steps) and fine-grained details (late clean steps). We aggregate features from the final 75\% of denoising steps using linearly increasing weights from 0.1 to 1.0, assigning higher importance to features extracted from less noisy intermediate states. Specifically, for timesteps $t \in [T/4, 0]$, we compute weighted features $\mathbf{F}_v^{\text{diff}} = \sum_{t=0}^{3T/4} w_t \mathbf{F}_v^{(t)}$ where weights $w_t$ increase linearly, yielding view-specific diffusion features that are normalized to unit length.

Diffusion features alone lack fine-grained spatial correspondence. We therefore fuse them with DINOv2\cite{oquab2023dinov2} features $\mathbf{F}_v^{\text{dino}}$ (768-dimensional from ViT-B/14) extracted from the textured rendering using equation~\ref{eq:1}, which provide strong local matching cues. The concatenated features
\begin{equation}
    \mathbf{F}_v^{\text{fused}} = [\alpha \mathbf{F}_v^{\text{diff}}; (1-\alpha) \mathbf{F}_v^{\text{dino}}]
    \label{eq:1}
\end{equation}

with $\alpha = 0.5$ are renormalized, producing 2048-dimensional per-pixel descriptors (1280 from diffusion + 768 from DINO). To transfer these 2D features to 3D, we define an unprojection operator $\mathbf{U}_v$ that maps each vertex $\mathbf{v}_i$ to its corresponding pixel location using known camera parameters. For robustness against projection noise, we apply ball query aggregation with $K = 100$ neighbors, each vertex averages features from all neighbors within radius $r$ (1\% of bounding box diagonal), promoting local consensus. Finally, we aggregate across all views via averaging with the help of equation ~\ref{eq:2}
\begin{equation}
    \mathbf{s}_i = \frac{1}{V} \sum_{v=1}^{V} \mathbf{U}_v(\mathbf{F}_v^{\text{fused}})_i
\label{eq:2}
\end{equation}

producing per-vertex semantic descriptors $\mathbf{S} = \{\mathbf{s}_1, \ldots, \mathbf{s}_N\} \in \mathbb{R}^{N \times 2048}$.

\paragraph{Design rationale:} Multi-view rendering is essential because single views suffer from occlusion—salient regions on the back of an object would receive no semantic signal. Geometry conditioning (depth and normals) ensures the diffusion model respects 3D structure rather than hallucinating textures that contradict the surface. The emphasis on later denoising timesteps prioritizes well-formed semantic features over early coarse abstractions. This entire pipeline runs offline once per mesh, treating diffusion as a frozen perceptual prior rather than a trainable component.

\subsection{Geometric Encoding and Efficient Sampling} 

Operating directly on high-resolution meshes (often $N > 50{,}000$ vertices) is computationally prohibitive for modern point cloud networks. We uniformly sample $M = 2{,}048$ points $\mathbf{P} \subset \mathbf{V}$ from the full vertex set, gathering corresponding normals $\tilde{\mathbf{N}}$ and semantics $\tilde{\mathbf{S}}$ at these indices. The geometric descriptors $\tilde{\mathbf{G}} = [\mathbf{P}; \tilde{\mathbf{N}}] \in \mathbb{R}^{M \times 6}$ encode both position and local surface orientation.

We encode these geometric descriptors using Point Transformer V3 as our backbone. This architecture serializes irregular point clouds into structured sequences via voxelization and space-filling curves (Z-order or Hilbert), enabling efficient patch-based self-attention. The model hierarchically abstracts spatial information through four encoder stages with progressively increasing channels (32, 64, 128, 256) and spatial downsampling, followed by a three-stage decoder (64, 128, 256 channels) with skip connections. We do not modify the architecture; we simply leverage its ability to capture multi-scale geometric context. The backbone outputs 64-dimensional features $\mathbf{H}_{\text{geo}}^{\text{raw}} \in \mathbb{R}^{M \times 64}$ due to encoder-decoder concatenation, which we project to a fixed hidden dimension $d_h = 32$ via $\mathbf{H}_{\text{geo}} = \mathbf{H}_{\text{geo}}^{\text{raw}} \mathbf{W}_{\text{geo}}$ followed by batch normalization and ReLU, yielding $\mathbf{H}_{\text{geo}} \in \mathbb{R}^{M \times 32}$.

The semantic stream applies a two-layer MLP bottleneck to compress the high-dimensional semantic features $\tilde{\mathbf{S}} \in \mathbb{R}^{M \times 2048}$ into the same 32-dimensional latent space with the help of the equation ~\ref{eq:3}.

\begin{equation}
    \mathbf{H}_{\text{sem}} = \text{ReLU}(\text{BN}(\mathbf{W}_2 \text{ReLU}(\text{BN}(\mathbf{W}_1 \tilde{\mathbf{S}}))))
    \label{eq:3}
\end{equation}

with intermediate dimension 512, producing $\mathbf{H}_{\text{sem}} \in \mathbb{R}^{M \times 32}$. This compression is necessary for computational efficiency and prevents overfitting to the very high-dimensional semantic space.

\subsection{Asymmetric Geometry-to-Semantics Fusion} 

The key architectural decision is how to combine geometric and semantic information. Naive element-wise addition or concatenation treats both modalities symmetrically, but this contradicts human attention theory. In biological vision, bottom-up geometric features (edges, motion, contrast) trigger initial orienting responses, which then retrieve top-down semantic knowledge ("is this a face?"). We formalize this asymmetry through cross-attention where geometry queries semantics.

We compute $\mathbf{Q} = \mathbf{H}_{\text{geo}} \mathbf{W}_Q$, $\mathbf{K} = \mathbf{H}_{\text{sem}} \mathbf{W}_K$, $\mathbf{V} = \mathbf{H}_{\text{sem}} \mathbf{W}_V$, and apply multi-head attention (4 heads) with head dimension $d_k = \frac{d_h}{4} = 8$ using equation ~\ref{eq:4}.

\begin{equation}
\mathbf{H}_{\text{attn}} = \text{MultiHead}(\mathbf{Q}, \mathbf{K}, \mathbf{V}) = \text{Concat}(\text{head}_1, \ldots, \text{head}_4) \mathbf{W}_O
\label{eq:4}
\end{equation}

where each head computes ${head}_\ell$ by equation ~\ref{eq:5}.
\begin{equation}
    \text{head}_\ell = \text{softmax}\left(\frac{\mathbf{Q}_\ell \mathbf{K}_\ell^\top}{\sqrt{d_k}}\right) \mathbf{V}_\ell
\label{eq:5}
\end{equation}
The geometric features query the semantic space, retrieving semantic content relevant to each geometric location. We fuse via residual connection $\mathbf{H}_{\text{fused}} = \mathbf{H}_{\text{geo}} + \mathbf{H}_{\text{attn}}$, ensuring geometry is never discarded. A two-layer MLP head (32 → 64 → 1) with sigmoid activation produces saliency predictions, which formulated on equation ~\ref{eq:6},
\begin{equation}
    \hat{y}_i^{(M)} = \sigma(\mathbf{w}_2^\top \text{ReLU}(\mathbf{W}_1 \mathbf{h}_{\text{fused},i}))
\label{eq:6}
\end{equation}
for each of the $M$ sampled points.

\paragraph{Why this asymmetry matters:} If semantics queried geometry, the model could ignore geometric anomalies when semantics are strong, causing it to miss geometrically salient but semantically boring regions (e.g., a sharp edge on a uniform surface). Our design ensures geometry always has "veto power"—even semantically important regions must be geometrically distinctive to attract attention. Ablations confirm this: reversing the attention direction degrades performance on meshes with high geometric complexity.

\subsection{Training with Hybrid Loss} 

Since predictions are at resolution $M = 2{,}048$ but ground truth is at full resolution $N$, we interpolate predictions to the full mesh via inverse distance weighting with $K = 3$ nearest neighbors with equation ~\ref{eq:7}

\begin{equation}
    \hat{y}_i^{(N)} = \sum_{j \in \mathcal{N}_3(\mathbf{v}_i)} w_{ij} \hat{y}_j^{(M)} / \sum_j w_{ij}
\label{eq:7}
\end{equation}
where $w_{ij} = \frac{1}{d_{ij} + \epsilon} $ with $\epsilon = 10^{-8}$, and $\mathcal{N}_3$ denotes the 3 nearest sampled points. We optimize a hybrid loss combining KL divergence for distributional match and correlation coefficient for spatial structure preservation with equation ~\ref{eq8}

\begin{equation}
    \mathcal{L} = 10 \cdot \mathcal{L}_{\text{KL}} - 2 \cdot \mathcal{L}_{\text{CC}}
    \label{eq8}
\end{equation}

where the negative sign on correlation reflects that higher correlation values indicate better alignment. The weight on KL divergence emphasizes distributional accuracy, while the correlation term ensures the spatial pattern of saliency is preserved. These metrics are standard in the saliency literature; KL divergence measures information loss when approximating the true distribution, and linear correlation coefficient captures the strength of spatial correspondence between predicted and ground truth heatmaps.

\subsection{Scanpath Generation via Reinforcement Learning} 

Static saliency maps predict where humans look but not when or in what order. To model temporal attention dynamics, we frame scanpath generation as a POMDP $\langle \mathcal{S}, \mathcal{A}, \mathcal{T}, \mathcal{R}, \mathcal{O}, \Omega \rangle$. The state $s_t = (v_t, \mathbf{m}_t)$ consists of the current vertex $v_t$ and a visit count vector $\mathbf{m}_t \in \mathbb{N}^N$ tracking how many times each vertex has been visited. At each step, the agent observes the $K = 6$ nearest neighbors on the mesh surface and selects one to move to, defining a discrete action space $\mathcal{A} = \{1, \ldots, K\}$ that respects mesh topology.

The reward function balances multiple objectives to encourage human-like scanpath behavior using equation ~\ref{eq:9}.

\begin{equation}
\begin{split}
    \mathcal{R}(s_t, a_t) = &\hat{y}_{v_{t+1}}^{(N)} - 0.2 \cdot \tanh\left(\frac{\mathbf{m}_t(v_{t+1})}{2}\right) \\
    &+ 0.15 \cdot \mathbb{I}[\text{new\_region}(v_{t+1})] \\
    &+ 0.1 \cdot \text{diversity}(v_{t+1}) - 0.05
\end{split}
\label{eq:9}
\end{equation}

The first term rewards moving to high-saliency regions using frozen SemGeo-AttentionNet predictions. The second term implements inhibition of return through a tanh-based penalty that saturates for frequently visited locations, with coefficient 0.2. The third term provides an exploration bonus (0.15) when entering previously unvisited spatial regions. The fourth term encourages diversity by rewarding movements that maintain distance from recent fixations (0.1). The final constant (-0.05) provides a small step penalty to discourage excessively long fixation sequences. This multi-component reward structure encourages the agent to explore salient regions while avoiding redundant revisits.

The observation $\mathbf{o}_t \in \mathbb{R}^{16}$ includes current saliency, current IOR value, saliency values at the 6 neighboring candidate vertices, Euclidean distances to these neighbors, episode progress, and distance to the mesh center, providing rich contextual information for action selection.

We train a policy $\pi_\phi: \mathcal{O} \rightarrow \Delta(\mathcal{A})$ using Proximal Policy Optimization (PPO) with standard hyperparameters (clip $\epsilon = 0.2$, GAE $\lambda = 0.95$, discount $\gamma = 0.99$). Episodes terminate after $T = 20$ fixations, producing scanpaths of length 20. The trained policy generates human-like exploration sequences that balance saliency-driven attention with spatial exploration, capturing both the "what" (salient regions) and "how" (temporal dynamics) of visual attention.

\paragraph{Why RL on meshes:} Existing scanpath models operate on 2D images with pixel-based actions. Our formulation respects 3D surface topology, actions navigate the mesh graph rather than a Euclidean grid. This is essential because attention on 3D objects depends on surface connectivity, not Euclidean distance (e.g., opposite sides of a thin object are spatially close but topologically distant).

\section{Experiments}
\label{sec:experiments}
\begin{figure*}[htbp]
  \centering
  \includegraphics[width=0.8\textwidth]{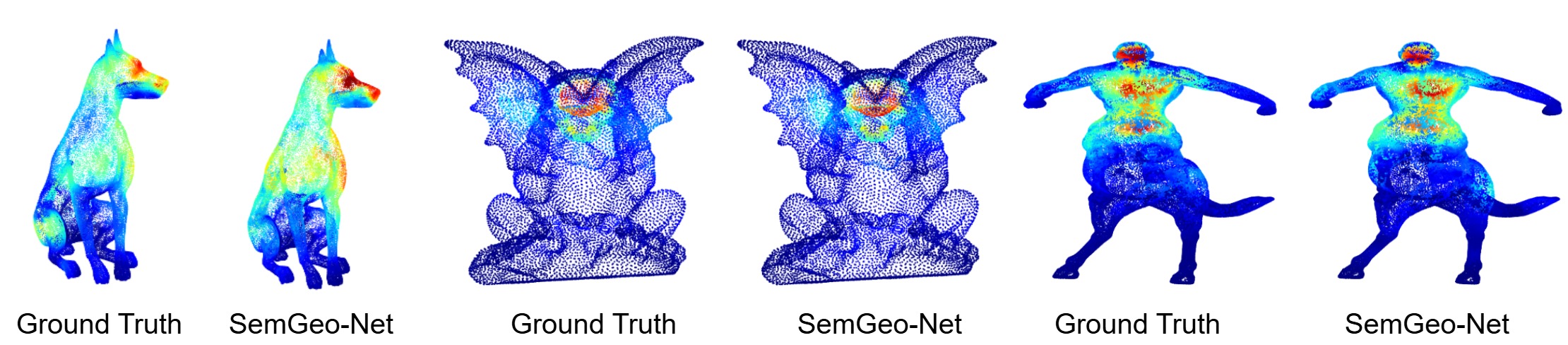} 
  \caption{\textbf{Results Gallery.} SemGeoAttentionNet's performance on Sal3D dataset. Warm areas are the salient part.}
  \label{fig:dataset1}
\end{figure*}

\begin{figure*}[htbp]
  \centering
  \includegraphics[width=0.8\textwidth]{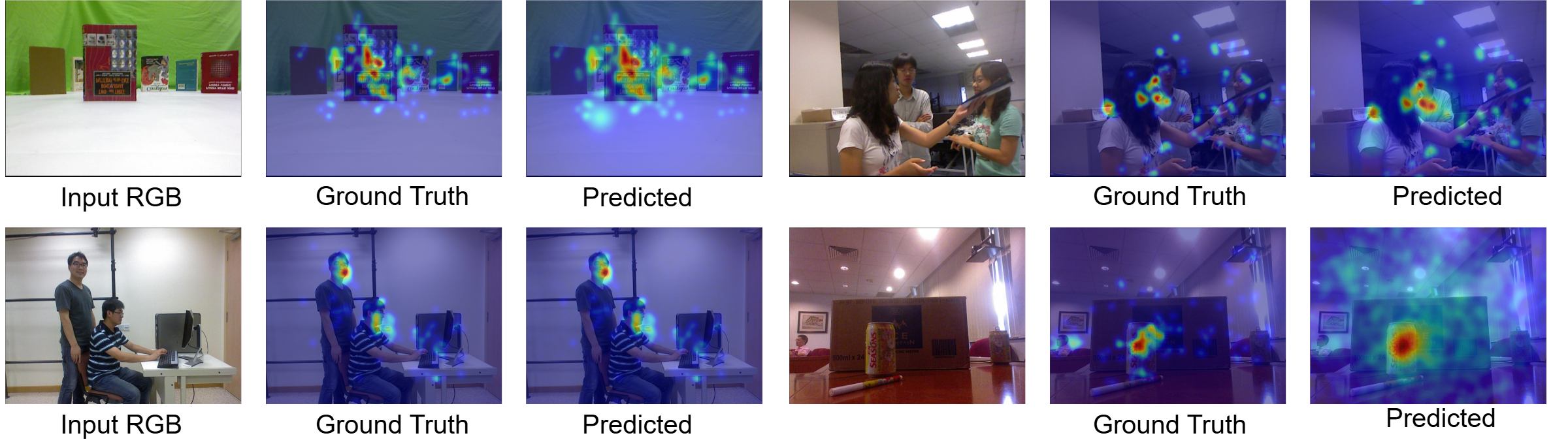} 
  \caption{\textbf{Results Gallery.} SemGeoAttentionNet's performance on NUS3D dataset. Warm represents the salient part and the comparision between GT and Prediction shows the accurateness of our model.}
  \label{fig:dataset2}
\end{figure*}


In this section we have discussed the quantitative and qualitative analysis of our proposed method with comparison to the prior works. Also we have discussed the datass

\subsection{Setup}
\label{subsec:setup}




\paragraph{Offline Preprocessing.} We extract semantic features through geometry-conditioned diffusion rendering. Each mesh is rendered from 100 viewpoints at 512×512 resolution using Stable Diffusion v1.5 with dual ControlNets, yielding 2048-dimensional per-vertex semantic descriptors.

\paragraph{Saliency Model Training.} We train SemGeo-AttentionNet on 2 NVIDIA A100 GPUs using AdamW optimizer (learning rate 1e-4) for 100 epochs with batch size 8. The hybrid loss combines 10×KL divergence and -2×correlation coefficient.

\paragraph{RL Scanpath Training.} We train the scanpath policy using Proximal Policy Optimization over 200,000 timesteps. Episodes consist of 20 fixations with discrete actions over 6-nearest mesh neighbors, balanced by saliency reward and IOR penalty ($\beta$=0.2).

\subsection{Dataset}
\label{subsec:data}
We evaluate SemGeo-AttentionNet on three established 3D visual attention datasets: SAL3D\cite{martin2024sal3d}, NUS3D-Saliency\cite{lang2012depth}, and 3DVA\cite{zhou2022edvam}, comparing against state-of-the-art geometric and learning-based methods across multiple metrics. Our approach demonstrates substantial improvements in both distributional accuracy and spatial structure preservation, validating the effectiveness of asymmetric geometry-semantics fusion for modeling human visual attention on 3D surfaces.

\subsection{Evaluation Metrics}
\label{subsec:eval}
We assess model performance using established metrics from the visual attention literature that capture different aspects of saliency prediction quality: distributional similarity, spatial structure preservation, and fixation-level accuracy.

\paragraph{Kullback-Leibler Divergence (KL-Div).} KL divergence measures the information loss when using the predicted saliency distribution to approximate the ground truth distribution. Given normalized saliency maps $\mathbf{y} = \{y_1, \ldots, y_N\}$ (ground truth) and $\hat{\mathbf{y}} = \{\hat{y}_1, \ldots, \hat{y}_N\}$ (prediction):

\begin{equation}
\text{KL}(\mathbf{y} \| \hat{\mathbf{y}}) = \sum_{i=1}^{N} y_i \log\left(\frac{y_i}{\hat{y}_i + \epsilon}\right)
\end{equation}

where $\epsilon = 10^{-8}$ prevents numerical instability. Lower values indicate better distributional match.

\paragraph{Correlation Coefficient (CC).} Pearson correlation measures the linear relationship between predicted and ground truth saliency values:

\begin{equation}
\text{CC}(\mathbf{y}, \hat{\mathbf{y}}) = \frac{\sum_{i=1}^{N} (y_i - \bar{y})(\hat{y}_i - \bar{\hat{y}})}{\sqrt{\sum_{i=1}^{N} (y_i - \bar{y})^2} \sqrt{\sum_{i=1}^{N} (\hat{y}_i - \bar{\hat{y}})^2}}
\end{equation}

Higher positive values indicate stronger correlation, with perfect prediction yielding CC = 1.

\paragraph{Normalized Scanpath Saliency (NSS).} NSS evaluates prediction quality at human fixation locations $\mathcal{F} = \{f_1, \ldots, f_F\}$:

\begin{equation}
\text{NSS}(\hat{\mathbf{y}}, \mathcal{F}) = \frac{1}{F} \sum_{j=1}^{F} \frac{\hat{y}_{f_j} - \mu_{\hat{y}}}{\sigma_{\hat{y}}}
\end{equation}

Values above 1.0 indicate above-chance performance, with values above 2.0 considered excellent.

\paragraph{Area Under ROC Curve (AUC-Judd).} AUC measures the model's ability to discriminate between fixated and non-fixated locations. AUC ranges from 0 to 1, with 0.5 indicating chance performance and higher values indicating better ranking of salient regions.

\paragraph{MultiMatch.} For scanpath evaluation, MultiMatch quantifies similarity between predicted and ground truth scanpath sequences across five dimensions: shape, direction, length, position, and duration. Scores range from 0 (completely dissimilar) to 1 (identical), with higher values indicating more human-like temporal attention patterns.

These complementary metrics provide comprehensive assessment: KL measures distributional accuracy, CC captures spatial structure, NSS evaluates fixation-level precision, AUC assesses discriminative power, and MultiMatch quantifies temporal scanpath similarity.

\subsection{Results}
\label{subsec:results}
\paragraph{SAL3D Dataset.} Table 1 presents results on the SAL3D benchmark, which comprises 58 meshes with eye-tracking data from 32 participants in VR. Our model achieves CC of 0.8492, a 28\% improvement over the previous best result (0.6616). KL divergence decreases to 0.1638 from 0.3051, while MSE drops to 0.0114 from 0.0204, demonstrating that semantic priors effectively complement geometric processing.

\begin{table}[h]
\centering
\caption{Quantitative comparison on SAL3D dataset.}
\begin{tabular}{lccc}
\toprule
\textbf{Method} & \textbf{CC} $\uparrow$ & \textbf{KL-Div} $\downarrow$ & \textbf{MSE} $\downarrow$ \\
\midrule
Song et al.\cite{song2019mesh} & 0.1249 & 0.7034 & 0.3220 \\
Nousias et al.\cite{nousias2020mesh} & 0.0570 & 1.9618 & 0.0759 \\
SAL3D model\cite{martin2024sal3d} & 0.6616 & 0.3051 & 0.0204 \\
Mesh Mamba\cite{zhang2025mesh} & 0.6140 & 0.3067 & - \\
\textbf{Ours (SemGeo-Attn)} & \textbf{0.8492} & \textbf{0.1638} & \textbf{0.0114} \\
\bottomrule
\end{tabular}
\end{table}

\paragraph{NUS3D-Saliency Dataset.} On NUS3D, our model achieves LCC of 0.609 and AUC-Judd of 0.935 (Table 2), improving over MIMO-GAN by 129\% in LCC and 22\% in AUC. The substantial AUC improvement indicates correct ranking of salient regions with high precision.

\begin{table}[h]
\centering
\caption{Quantitative comparison on NUS3D-Saliency dataset.}
\begin{tabular}{lcc}
\toprule
\textbf{Method} & \textbf{LCC} $\uparrow$ & \textbf{AUC} $\uparrow$ \\
\midrule
DSM & 0.222 & 0.726 \\
MIMO-GAN-A1 & 0.290 & 0.781 \\
MIMO-GAN-A2 & 0.057 & 0.584 \\
MIMO-GAN-A3 & 0.259 & 0.753 \\
MIMO-GAN & 0.267 & 0.761 \\
\textbf{Ours (SemGeo-Attn)} & \textbf{0.609} & \textbf{0.935} \\
\bottomrule
\end{tabular}
\end{table}

\paragraph{3DVA Dataset.} On 3DVA, our model achieves mean LCC of 0.762 with standard deviation 0.093 (Table 3), outperforming MIMO-GAN-CRF (0.510 LCC) by 49\%. The reduced variance indicates consistent prediction quality across diverse stimuli.

\begin{table}[h]
\centering
\caption{Quantitative comparison on 3DVA dataset.}
\begin{tabular}{lcc}
\toprule
\textbf{Method} & \textbf{Mean LCC} $\uparrow$ & \textbf{Std. Dev.} $\downarrow$ \\
\midrule
Multi-Scale Gaussian & 0.131 & 0.265 \\
Diffusion Wavelets & 0.088 & 0.222 \\
Spectral Processing & 0.078 & 0.253 \\
Point Clustering & 0.132 & 0.300 \\
Salient Regions & 0.215 & 0.245 \\
Hilbert-CNN & 0.113 & 0.267 \\
RPCA & 0.199 & 0.251 \\
CfS-CNN & 0.226 & 0.243 \\
MIMO-GAN-CRF & 0.510 & 0.108 \\
\textbf{Ours (SemGeo-Attn)} & \textbf{0.762} & \textbf{0.093} \\
\bottomrule
\end{tabular}
\end{table}

\paragraph{Fixation-Level Accuracy.} Our model achieves NSS of 2.05 on SAL3D and 1.60 on NUS3D (Table 4). Values above 2.0 indicate excellent performance, confirming that predictions capture both global patterns and precise fixation locations.


\subsection{Scanpath Generation Results} 

The RL-based scanpath generator achieves scanpath NSS of 2.05 and MultiMatch score of 0.51 on NUS3D, indicating temporal fixation sequences closely match human viewing behavior. The policy successfully balances saliency-driven attention with spatial exploration and inhibition of return.

\subsection{Qualitative Analysis}
\label{subsec:quali}
\begin{figure}[t]
  \centering
  \includegraphics[width=0.5\textwidth ]{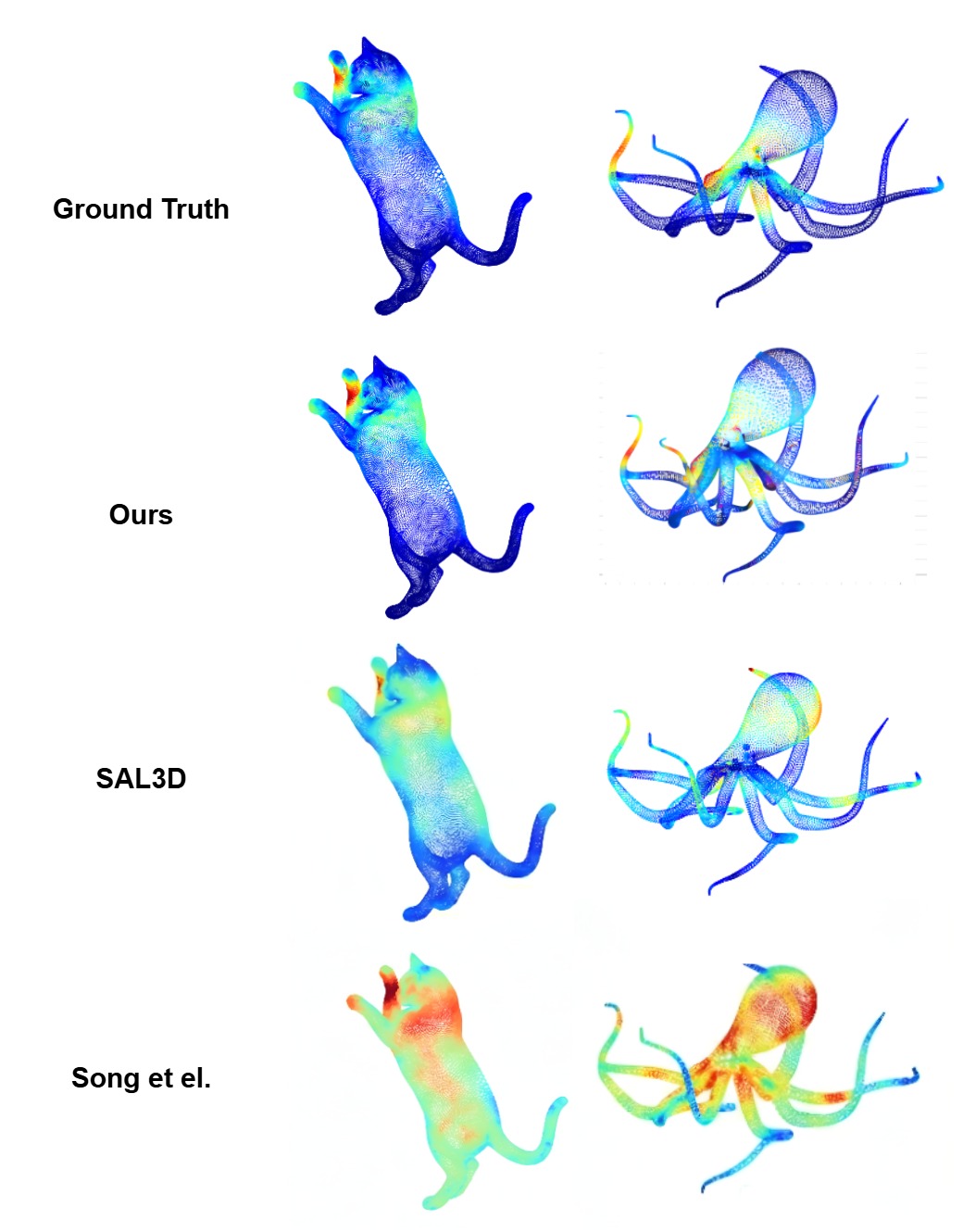} 
  \caption{\textbf{Comparison.}. Our model's performance on Sal3D along side benchmark model \cite{martin2024sal3d} and \cite{song2019mesh}  }
  \label{fig:quali}
\end{figure}
Figure ~\ref{fig:quali} presents qualitative comparisons on representative meshes from SAL3D dataset. On the octopus mesh from SAL3D, our model correctly predicts high saliency on the head and tentacle curvatures, matching ground truth fixations, while geometric-only methods over-emphasize uniform surface bumps. The gargoyle face demonstrates semantic grounding: despite being geometrically smooth, facial features receive high predicted saliency matching human fixations, a pattern that geometry-based methods fail to capture. These examples validate our hypothesis that bottom-up geometric processing must be modulated by top-down semantic recognition to accurately model human visual attention on 3D surfaces.


\section{Conclusion}
\label{sec:conclusion}

This article proposed SemGeo-AttentionNet, a dual-stream framework for modeling human visual attention on three-dimensional surfaces through explicit geometry-semantics integration. The architecture combines Point Transformer V3 for geometric processing with diffusion-based semantic priors from geometry-conditioned multi-view rendering, unified through asymmetric cross-attention where geometric features query semantic content. This design formalizes the cognitive dichotomy underlying human visual attention, ensuring bottom-up distinctiveness guides top-down retrieval. The framework extends to temporal scanpath prediction through reinforcement learning with surface-aware action spaces respecting mesh topology and inhibition-of-return dynamics. Experimental validation on SAL3D, NUS3D, and 3DVA benchmarks demonstrates state-of-the-art performance, establishing that cognitively motivated architectures can effectively capture human visual attention mechanisms on three-dimensional objects.

\bibliographystyle{unsrt}  
\bibliography{references}

\newpage
\appendix
\onecolumn
\clearpage
\section*{Supplementary Material}

\section{Implementation Details}

\paragraph{Rendering.}
We implement multi-view rendering using PyTorch3D with Hard Phong shading. Each mesh is rendered from 100 uniformly distributed viewpoints at 512$\times$512 resolution. We sample viewpoints by varying elevation ($\theta$) and azimuthal ($\phi$) angles in a 10$\times$10 grid with linearly spaced intervals between 0$^\circ$--360$^\circ$. Point lights are positioned at camera locations with ambient color (0.6, 0.6, 0.6), diffuse color (0.4, 0.4, 0.5), and minimal specular contribution (0.01, 0.01, 0.01). The camera distance is set to 0.65$\times$ the mesh bounding box diagonal to ensure complete object visibility across all poses.

\paragraph{Diffusion.}
We employ Stable Diffusion v1.5 with dual ControlNet conditioning for geometry-aware texture synthesis. The depth ControlNet (\texttt{control\_v11f1p\_sd15\_depth}) and normal ControlNet (\texttt{control\_v11p\_sd15\_normalbae}) jointly condition the generation process. We use 30 DDIM inference steps with guidance scale 7.0 and $\eta=1.0$. Text prompts follow the template: ``\texttt{\{object\}, best quality, highly detailed, photorealistic, photo}'' with negative prompt ``\texttt{lowres, low quality, monochrome, watermark}'' to ensure high-fidelity textured renderings.

\paragraph{Feature Extraction.}
Semantic descriptors combine DINOv2 ViT-B/14 features (768-dim) with Stable Diffusion U-Net decoder features (1280-dim), yielding 2048-dimensional per-vertex embeddings. For each rendered view, we extract DINO features at 518$\times$518 resolution (37$\times$37 spatial tokens) and diffusion features from the U-Net bottleneck. Features are mapped to mesh vertices through ball query aggregation with $K=100$ nearest pixels and 1\% radius tolerance relative to mesh extent. Missing vertex features are filled via nearest-neighbor propagation.

\paragraph{Network Architecture.}
SemGeo-AttentionNet employs a two-stream architecture with cross-attention fusion:

\textbf{Geometry Stream:} Point Transformer V3 (PTv3) processes 6-channel input (XYZ coordinates + vertex normals). The encoder uses depths (2, 2, 6, 2) with channels (32, 64, 128, 256) and attention heads (2, 4, 8, 16). The decoder mirrors with depths (1, 1, 1), channels (64, 128, 256), and heads (4, 8, 16). We use patch size 48, MLP ratio 4, and drop path rate 0.3.

\textbf{Semantic Stream:} A two-layer MLP projects 2048-dim semantic features through 512 hidden units to 32-dim embeddings with BatchNorm and ReLU activations.

\textbf{Fusion:} 4-head cross-attention with geometry embeddings as queries and semantic embeddings as keys/values. Residual connection adds attention output to geometry features.

\textbf{Prediction Head:} MLP (32$\rightarrow$64$\rightarrow$1) with Sigmoid activation outputs per-vertex saliency scores in $[0,1]$.

\section{Compute Time}

\paragraph{Offline Preprocessing.}
We extract semantic features through geometry-conditioned diffusion rendering as a one-time preprocessing step. Each mesh is rendered from 100 uniformly sampled viewpoints at 512$\times$512 resolution. We employ Stable Diffusion v1.5 with dual ControlNets (depth and normal conditioning) using 30 inference steps and guidance scale 7. Features from the U-Net decoder (1280-dim) are fused with DINOv2 ViT-B/14 features (768-dim), yielding 2048-dimensional per-vertex semantic descriptors through multi-view unprojection with 1\% ball query radius.

\paragraph{Saliency Model Training.}
We train SemGeo-AttentionNet on single NVIDIA A100 GPUs with 80GB memory using AdamW optimizer (learning rate 1e-4, weight decay 1e-4) for 100 epochs with batch size 8. The network processes 2048 uniformly sampled points with hidden dimension 32 and 4-head cross-attention. Training completes in approximately 4 hours for SAL3D and 3.5 hours for NUS3D. The hybrid loss combines 10$\times$KL divergence and $-2\times$correlation coefficient, with predictions interpolated to full resolution via inverse distance weighting ($K=3$ neighbors).

\paragraph{RL Scanpath Training.}
We train the scanpath policy using Proximal Policy Optimization with standard hyperparameters (clip $\epsilon=0.2$, GAE $\lambda=0.95$, discount $\gamma=0.99$) over 200,000 timesteps. Episodes consist of 20 fixations with discrete actions over 6-nearest mesh neighbors, balanced by saliency reward and IOR penalty ($\beta=0.2$).

\section{Ablation Studies}

\subsection{Effect of Number of Views}
We ablate the number of rendered viewpoints used for semantic feature extraction in Table~\ref{table:views-ablation}. 

\begin{table}[h]
\caption{\textbf{Ablation on number of rendered views.} Performance on SAL3D test set. Increasing views improves feature coverage but with diminishing returns beyond 100 views.}
\centering
\begin{tabular}{c|cc}
\toprule
\textbf{Num Views} & \textbf{KLD} $\downarrow$ & \textbf{CC} $\uparrow$ \\
\midrule
25 & 0.2847 & 0.7126 \\
50 & 0.2103 & 0.7854 \\
100 (Ours) & 0.1638 & 0.8492 \\
200 & 0.1592 & 0.8531 \\
\bottomrule
\end{tabular}
\label{table:views-ablation}
\end{table}

\subsection{Effect of Feature Components}
We evaluate the individual contributions of DINO and Stable Diffusion features in Table~\ref{table:features-ablation}.

\begin{table}[h]
\caption{\textbf{Ablation on semantic feature components.} Combining DINO's semantic understanding with diffusion's texture-aware features yields optimal performance.}
\centering
\begin{tabular}{l|cc}
\toprule
\textbf{Features} & \textbf{KLD} $\downarrow$ & \textbf{CC} $\uparrow$ \\
\midrule
Geometry only (no semantic) & 0.3892 & 0.5847 \\
DINO only (768-dim) & 0.2314 & 0.7623 \\
SD only (1280-dim) & 0.2567 & 0.7281 \\
Combined (2048-dim, Ours) & 0.1638 & 0.8492 \\
\bottomrule
\end{tabular}
\label{table:features-ablation}
\end{table}

\subsection{Effect of Fusion Method}
We compare different strategies for fusing geometric and semantic streams in Table~\ref{table:fusion-ablation}.

\begin{table}[h]
\caption{\textbf{Ablation on fusion strategies.} Cross-attention allows geometry to selectively attend to relevant semantic information.}
\centering
\begin{tabular}{l|cc}
\toprule
\textbf{Fusion Method} & \textbf{KLD} $\downarrow$ & \textbf{CC} $\uparrow$ \\
\midrule
Concatenation & 0.2156 & 0.7834 \\
Element-wise Addition & 0.2287 & 0.7692 \\
Self-Attention & 0.1923 & 0.8147 \\
Cross-Attention (Ours) & 0.1638 & 0.8492 \\
\bottomrule
\end{tabular}
\label{table:fusion-ablation}
\end{table}

\subsection{Effect of Loss Weights}
We ablate the weighting of KL divergence and correlation coefficient in our hybrid loss function in Table~\ref{table:loss-ablation}.

\begin{table}[h]
\caption{\textbf{Ablation on loss function weights.} Hybrid loss: $\mathcal{L} = \alpha \cdot \text{KLD} - \beta \cdot \text{CC}$.}
\centering
\begin{tabular}{cc|cc}
\toprule
$\alpha$ (KLD) & $\beta$ (CC) & \textbf{KLD} $\downarrow$ & \textbf{CC} $\uparrow$ \\
\midrule
1 & 0 & 0.1847 & 0.8056 \\
10 & 0 & 0.1623 & 0.8194 \\
1 & 2 & 0.2134 & 0.8312 \\
10 & 2 (Ours) & 0.1638 & 0.8492 \\
\bottomrule
\end{tabular}
\label{table:loss-ablation}
\end{table}

\subsection{Effect of Sampled Points}
We ablate the number of points sampled during training in Table~\ref{table:points-ablation}.

\begin{table}[h]
\caption{\textbf{Ablation on number of sampled points.} More points improve spatial coverage but increase memory requirements.}
\centering
\begin{tabular}{c|cc}
\toprule
\textbf{Sampled Points} & \textbf{KLD} $\downarrow$ & \textbf{CC} $\uparrow$ \\
\midrule
512 & 0.2478 & 0.7512 \\
1024 & 0.1956 & 0.8078 \\
2048 (Ours) & 0.1638 & 0.8492 \\
4096 & 0.1587 & 0.8548 \\
\bottomrule
\end{tabular}
\label{table:points-ablation}
\end{table}

\subsection{Effect of Hidden Dimension}
We ablate the hidden dimension used in the fusion module in Table~\ref{table:hidden-ablation}.

\begin{table}[h]
\caption{\textbf{Ablation on hidden dimension.} Smaller dimensions reduce parameters while maintaining performance.}
\centering
\begin{tabular}{c|ccc}
\toprule
\textbf{Hidden Dim} & \textbf{KLD} $\downarrow$ & \textbf{CC} $\uparrow$ & \textbf{Params (M)} \\
\midrule
16 & 0.1812 & 0.8267 & 2.41 \\
32 (Ours) & 0.1638 & 0.8492 & 2.58 \\
64 & 0.1594 & 0.8536 & 3.12 \\
128 & 0.1601 & 0.8519 & 4.67 \\
\bottomrule
\end{tabular}
\label{table:hidden-ablation}
\end{table}

\section{Hyperparameter Summary}

\begin{table}[h]
\caption{\textbf{Complete hyperparameter settings.}}
\centering
\resizebox{\linewidth}{!}{
\begin{tabular}{l|l|l}
\toprule
\textbf{Component} & \textbf{Parameter} & \textbf{Value} \\
\midrule
\multirow{4}{*}{Rendering} & Number of views & 100 \\
& Resolution & 512$\times$512 \\
& Camera distance scale & 0.65 \\
& Lighting (ambient/diffuse/specular) & (0.6, 0.6, 0.6) / (0.4, 0.4, 0.5) / (0.01, 0.01, 0.01) \\
\midrule
\multirow{5}{*}{Diffusion} & Base model & Stable Diffusion v1.5 \\
& Depth ControlNet & control\_v11f1p\_sd15\_depth \\
& Normal ControlNet & control\_v11p\_sd15\_normalbae \\
& Inference steps & 30 \\
& Guidance scale / DDIM $\eta$ & 7.0 / 1.0 \\
\midrule
\multirow{5}{*}{Features} & DINO model & DINOv2 ViT-B/14 \\
& DINO dimension & 768 \\
& SD U-Net dimension & 1280 \\
& Total dimension & 2048 \\
& Ball query ($K$ / radius) & 100 / 1\% \\
\midrule
\multirow{7}{*}{PTv3 Encoder} & Input channels & 6 (XYZ + normals) \\
& Depths & (2, 2, 6, 2) \\
& Channels & (32, 64, 128, 256) \\
& Attention heads & (2, 4, 8, 16) \\
& Patch size & 48 \\
& MLP ratio & 4 \\
& Drop path & 0.3 \\
\midrule
\multirow{3}{*}{PTv3 Decoder} & Depths & (1, 1, 1) \\
& Channels & (64, 128, 256) \\
& Attention heads & (4, 8, 16) \\
\midrule
\multirow{3}{*}{Semantic MLP} & Layer 1 & 2048 $\rightarrow$ 512 \\
& Layer 2 & 512 $\rightarrow$ 32 \\
& Activations & BatchNorm + ReLU \\
\midrule
\multirow{3}{*}{Fusion} & Type & Cross-Attention \\
& Hidden dimension & 32 \\
& Attention heads & 4 \\
\midrule
\multirow{3}{*}{Prediction Head} & Layer 1 & 32 $\rightarrow$ 64 \\
& Layer 2 & 64 $\rightarrow$ 1 \\
& Output activation & Sigmoid \\
\bottomrule
\end{tabular}}
\label{table:hyperparams-model}
\end{table}

\begin{table}[h]
\caption{\textbf{Complete hyperparameter settings.}}
\centering
\resizebox{\linewidth}{!}{
\begin{tabular}{l|l|l}
\toprule
\textbf{Component} & \textbf{Parameter} & \textbf{Value} \\
\midrule
\multirow{6}{*}{Training} & Optimizer & AdamW \\
& Learning rate & 1e-4 \\
& Weight decay & 1e-4 \\
& Batch size & 8 \\
& Epochs & 100 \\
& Sampled points per mesh & 2048 \\
\midrule
\multirow{2}{*}{Loss Function} & KLD weight ($\alpha$) & 10 \\
& CC weight ($\beta$) & 2 \\
\midrule
\multirow{2}{*}{Interpolation} & Method & Inverse Distance Weighting \\
& $K$ neighbors & 3 \\
\midrule
\multirow{5}{*}{RL Scanpath} & Algorithm & PPO \\
& Clip $\epsilon$ & 0.2 \\
& GAE $\lambda$ / Discount $\gamma$ & 0.95 / 0.99 \\
& Timesteps & 200,000 \\
& IOR penalty $\beta$ & 0.2 \\
\bottomrule
\end{tabular}}
\label{table:hyperparams-training}
\end{table}






\end{document}